\documentclass[graybox, a4paper]{svmult}
\usepackage{helvet}         %
\usepackage{courier}        %
\usepackage{type1cm}        %
\usepackage{makeidx}         %
\usepackage{graphicx}        %
\usepackage{multicol}        %
\usepackage[bottom]{footmisc}%

\usepackage[T1]{fontenc}
\usepackage[hidelinks]{hyperref}
\usepackage{courier}
\usepackage{color}
\usepackage{upquote}
\usepackage{xcolor}
\usepackage{listings}
\usepackage{caption}
\usepackage[sort,compress]{cite}
\usepackage{paralist}

\usepackage[font=small,labelfont=bf]{caption}
\usepackage{bm}
\usepackage{mathtools}
\usepackage{subcaption}
\usepackage{graphicx, epstopdf}
\usepackage[framed, numbered]{matlab-prettifier}
\usepackage{flushend}
\usepackage{dsfont}
\usepackage{paralist}
\usepackage[nolist]{acronym}

\definecolor{hellgelb}{rgb}{1,1,0.85} 
\definecolor{colKeys}{rgb}{0,0,1} 
\definecolor{colIdentifier}{rgb}{0,0,0} 
\definecolor{colComments}{rgb}{0,0.5,0} 
\definecolor{colString}{rgb}{0.81,0.12,0.95}
\usepackage{amsmath}
\usepackage{amssymb}
\usepackage{placeins}

\newcommand{\tr}{^\top}

\newcommand{\defeq}{\vcentcolon=}

\newcommand{\repeatthanks}{\textsuperscript{\thefootnote}}

\begin{document}
\title*{\LARGE \bf LION: Lidar-Inertial Observability-Aware Navigator for Vision-Denied Environments
}
\titlerunning{LION: Lidar-Inertial Observability-Aware Navigator}
\author{Andrea Tagliabue\thanks{These authors contributed equally}$^{,1}$, Jesus Tordesillas\repeatthanks$^{,1}$, Xiaoyi Cai\repeatthanks$^{,1}$, Angel Santamaria-Navarro$^2$, Jonathan P. How$^1$, Luca Carlone$^1$, and Ali-akbar Agha-mohammadi$^2$}

\authorrunning{A. Tagliabue, J. Tordesillas,  X. Cai et al.}

\institute{$^1$ Massachusetts Institute of Technology, Cambridge, MA \email{{atagliab, jtorde, xyc, jhow, lcarlone}@mit.edu} \\
 $^{2}$ Jet Propulsion Laboratory, California Institute of Technology, Pasadena, CA
        \email{{angel.santamaria.navarro, aliagha}@jpl.nasa.gov}
This research work was partially carried out at the Jet Propulsion Laboratory, California Institute of Technology, under a contract with the National Aeronautics and Space Administration.
Government sponsorship acknowledged.
}
\maketitle

\vspace*{-1.0in}

\abstract{
State estimation for robots navigating in GPS-denied and perceptually-degraded environments, such as underground tunnels,  mines and planetary sub-surface voids~\cite{agha2019robotic}, remains challenging in robotics. Towards this goal, we present LION (Lidar-Inertial Observability-Aware Navigator), which is part of the state estimation framework developed by the team CoSTAR \cite{TeamCoST38:online} for the DARPA Subterranean Challenge \cite{DARPASub18:online}, where the team achieved second and first places in the Tunnel and Urban circuits in August 2019 and February 2020, respectively. 
LION provides high-rate odometry estimates by fusing high-frequency inertial data from an IMU and low-rate relative pose estimates from a lidar via a fixed-lag sliding window smoother. LION does not require knowledge of relative positioning between lidar and IMU, as the extrinsic calibration is estimated online. In addition, LION is able
to self-assess its performance using an observability metric that evaluates whether
the pose estimate is geometrically ill-constrained.
Odometry and confidence estimates are used by HeRO \cite{santamaria2019towards}, a supervisory algorithm that provides robust estimates by switching between different odometry sources.
In this paper we benchmark the performance of LION in perceptually-degraded subterranean environments, demonstrating its high technology readiness level for deployment in the field.
}
\noindent
\textbf{Video}: \textcolor{blue}{\href{https://youtu.be/Jd-sqBioarI}{https://youtu.be/Jd-sqBioarI}}
\vspace*{-.2in}
\begin{figure}[h]
\centering
\includegraphics[width=0.48\linewidth]{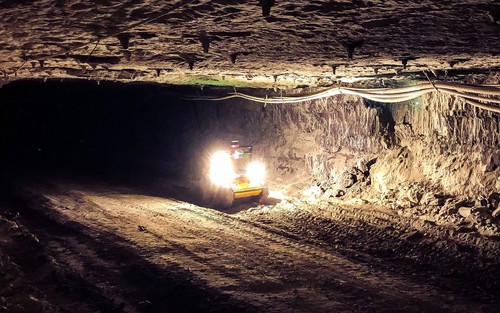}  
\includegraphics[width=0.48\linewidth]{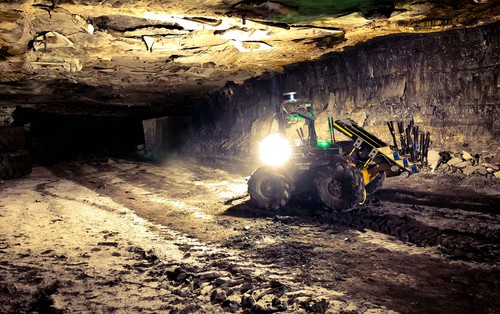}
  	\caption{Ground robots using LION to explore the Arch Coal Mine (West Virginia, USA), at approximately 275 meters (900 feet) underground.}
    \label{pic:mine}
    \vspace*{-.4in}
\end{figure}

\FloatBarrier
\section{INTRODUCTION}
Accurate and efficient odometry estimation for motion planning and control of robots in GPS-denied, perceptually-degraded, unknown underground environments still remains challenging in robotics. Common visual and visual-inertial solutions \cite{bloesch2015robust, qin2018vins, mur2015orb, forster2016svo} are not suitable for this task due to the potential absence of light or the strong illumination changes caused by mounting a light source on a moving robot. 
State estimation strategies for these environments \cite{neumann2014towards, papachristos2019autonomous} usually rely on proprioceptive sensing (e.g., IMU, wheel-encoders) and/or active exteroceptive methods (e.g., lidar, SoNaR, Time-of-Flight cameras or ultrasonic sensors). To compute odometry estimates in subterranean environments, the robots of the team CoSTAR relied on the complementarity of a lidar and an IMU as preferential sources of information.

Lidar odometry (LO) and Lidar-Inertial odometry (LIO) present multiple challenges, especially related with the trade-offs between estimation accuracy and computational complexity, and scene observability. Most of the current approaches compute lidar odometry via the Iterative-Closest Point (ICP) algorithm ~\cite{besl1992method, chen1992object, zhang1994iterative, magnusson2007scan, pomerleau2013comparing} or by extracting and matching features \cite{shan2018lego, ye2019tightly, zhang2014loam}. While the first approach is generally less computationally expensive, the second is more practical for the purpose of creating a sparse map used for efficient re-localization. In both cases, the odometry estimates can be refined by aligning lidar-scans or lidar-features with a local map \cite{zhang2014loam, shan2018lego}. Fusing the information from an IMU generally improves the estimation accuracy \cite{ye2019tightly, qin2019lins, hemann2016long, lin2019loam_livox}, since it further constrains the estimate and guarantees a high-output rate, thanks to techniques such as IMU pre-integration~\cite{forster2015imu}. Regardless of the scan-alignment technique used, lidar-based methods produce poor estimates in those scenarios that are not sufficiently geometrically-rich to constrain the motion estimation \cite{gelfand2003geometrically, zhang2016degeneracy, censi2007accurate}. %
This is especially true in tunnel-like environments \cite{neumann2014towards, papachristos2019autonomous}, where it is not easy to constrain the relative motion along the main shaft of the tunnel. Techniques employed to mitigate these issues consist of observability-aware point-cloud sampling \cite{gelfand2003geometrically}, degeneracy detection and mitigation \cite{zhang2016degeneracy, zhang2017enabling, hinduja2019degeneracy} and the fusion of other sources of information, such as inertial data from an IMU.  

This work presents LION, a Lidar-Inertial Observability-Aware algorithm for Navigation in perceptually-degraded environments, which 
was %
adopted by the team CoSTAR for the first year of the DARPA Subterranean Challenge. Our solution relies on the fusion of pre-integrated IMU measurements \cite{forster2015imu} with pose estimates from scan-to-scan ICP alignments, taking advantage of a fixed-lag smoother architecture \cite{kaess2012isam2, dellaert2012factor}. We simultaneously  calibrate online the extrinsics (translation and rotation) between the lidar and IMU sensors.
To address potential observability issues, we use a geometric observability score \cite{gelfand2003geometrically} that allows LION to predict potential degradation in its output given the geometric structure of the observed scene.
By reporting the proposed score, we can 
switch to a different state estimation algorithm (e.g., from wheel encoders, visual-inertial or thermal-inertial) via a supervisory algorithm, such as HeRO~\cite{santamaria2019towards}. This approach guarantees a continuous, reliable and gravity-aligned state estimate to the cascaded planning and control algorithms \cite{alatur2020autonomous}.

LION has been extensively tested in two coal mines of Beckley (West Virginia), two NIOSH experimental mines in Pittsburgh (Pennsylvania), a gold mine in Idaho Springs (Colorado), and many office-like environments at the Jet Propulsion Laboratory (California), showing its high readiness level for deployment in the field.

\vspace{-0.5cm}
\section{TECHNICAL APPROACH}\label{sec:LIO}
\begin{figure}[!tbp]
  \centering
  \begin{minipage}[b]{0.48\textwidth}
    \includegraphics[width=\textwidth]{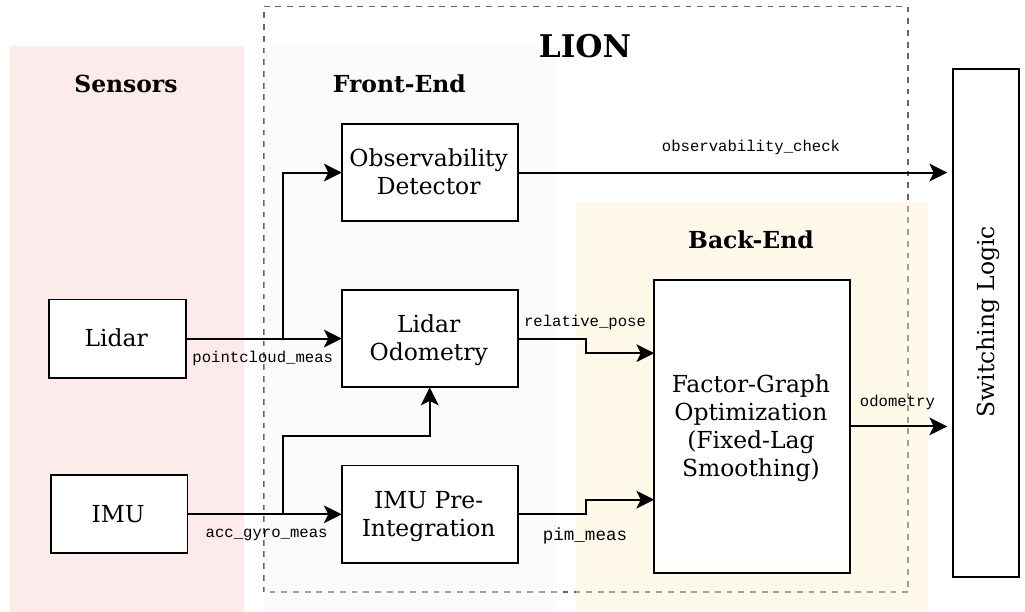}
    \caption{Front-end and back-end of LION. The front-end runs ICP and IMU pre-integration, whose outputs are used as factors in the back-end pose-graph optimization
    }
    \label{fig:diagram}
  \end{minipage}
  \hfill
  \begin{minipage}[b]{0.41\textwidth}
    \includegraphics[width=\textwidth]{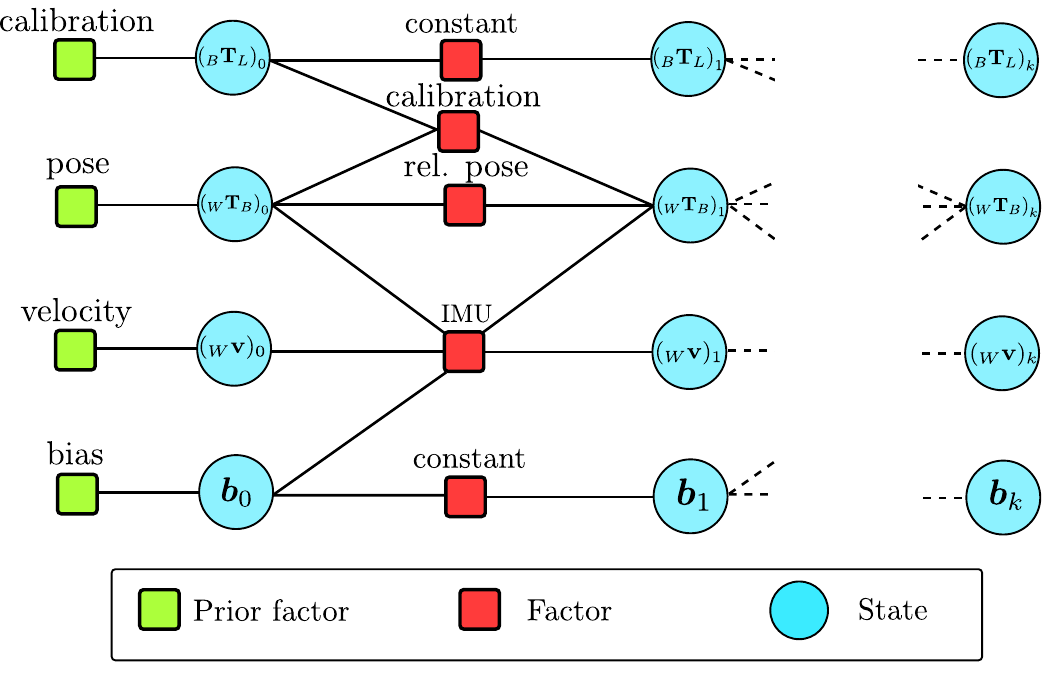}
    \caption{Factor graph representing state and measurements used by LION. In this figure we use the shortcut $\boldsymbol{b}_j:=\{\mathbf{b}^{a}, _{B}\mathbf{b}^{g}\}_j$ for the biases. 
    }
    \label{fig:factor_graph_colored}
  \end{minipage}
  \vspace*{-.2in}
\end{figure}

\textbf{Lidar-Inertial Odometry:}
 LION is a sliding window estimator 
 divided into two parts: a front-end consisting of the lidar-odometry and IMU pre-integration, and a factor-graph back-end (see Fig. \ref{fig:diagram}). In the following, we use an odometry reference frame $W$,
a body reference frame $B$, attached to the center of the IMU, and the lidar frame $L$, attached to the center of the lidar. The rigid transformation from a point expressed in a frame $A$ to a point expressed in frame $B$ is represented as the $4\times4$ matrix $_{B}\mathbf{T}_{A}$, which contains the rotation matrix $_{B}\mathbf{R}_{A}$ and translation vector $ _{B}\mathbf{t}_{A}$.

The \textit{front-end} of LION consists of three modules: lidar odometry, IMU pre-integration and the observability detector (see section \ref{sec:observability}). The lidar odometry module obtains relative pose measurements $ _{L_{k-1}}{\mathbf{T}}_{L_{k}}$ by using the Generalized Iterative Closest Point algorithm~\cite{segal2009generalized} on two lidar point clouds captured from consecutive lidar frames $L_{k}$ and $L_{k-1}$ at time stamps $k$ and $k-1$ respectively. To ease the convergence of the ICP algorithm, every incoming point cloud is pre-rotated in a gravity aligned-frame (i.e., setting the rotation estimated by the IMU as initial guess for ICP). The IMU pre-integration module leverages the state-of-the-art on-manifold pre-integration theory to summarize the high-rate IMU measurements into a single motion constraint~\cite{forster2015imu, forster2015manifold}. Alternatively, the scan-to-scan based front-end can be replaced by LOCUS \cite{palieri2020locus}, which additionally aligns the incoming scan to an incrementally built local map and performs a refinement step.

In the \textit{back-end}, the relative pose measurements produced by the front-end are stored in a factor graph in conjunction with the IMU measurements. A representation of the states and factors  used in the factor graph can be found in Fig. \ref{fig:factor_graph_colored}. 
Let us define the state vector $\mathbf{{x}}_{j}$ at the $j$-th time step as
$$\mathbf{{x}}_{j}:=\{\:_{W}\mathbf{T}_{B},\:_{W}\mathbf{v}
,\:_{B}\mathbf{b}^{a},\:_{B}\mathbf{b}^{g},\:_{B}\mathbf{T}_{L}\}_{j}\, 
$$ where $\:_{W}\mathbf{T}_{B}$ is the IMU-to-world transformation, $_W \mathbf{v}$ 
is the linear velocity, $_B \mathbf{b}^a$ and $_B \mathbf{b}^g$ are the accelerometer and gyroscope biases of the IMU, and $\:_{B}\mathbf{T}_{L}$ is the lidar-to-IMU transformation. Following the notation from \cite{forster2015manifold}, let $\mathcal{{K}}_{k}:=\{k-m+1,...,k\}$
denote the $m$ time steps inside the sliding window, and let $\mathcal{{X}}_{k}:=\{\mathbf{{x}}_{j}\}_{j\in\mathcal{{K}}_{k}}$ and $\mathcal{{Z}}_{k}$ denote respectively the states and measurements in this sliding window.
The factor graph optimization aims to solve therefore the following program \cite{forster2015manifold}: $
\mathcal{X}_{k}^{\star}:=\arg\min_{\mathcal{X}_{k}}\left(-\log_{e}p\left(\mathcal{X}_{k}|\mathcal{Z}_{k}\right)\right)
$, where $p\left(\mathcal{X}_{k}|\mathcal{Z}_{k}\right)$ is the posterior distribution. We model the optimization problem using GTSAM \cite{dellaert2012factor} and solve this optimization problem using iSAM2 \cite{kaess2012isam2}.

\label{sec:observa} \label{sec:observability}
\textbf{Observability Metric:} In subterranean environments, it is crucial for a state estimation algorithm to determine how well the geometry of the scene is able to constrain the estimation in all the translational directions. 
Following \cite{gelfand2003geometrically,bonnabel2016covariance}, and assuming that the rotation is small, the Hessian of the Point-To-Plane ICP cost is given by $2\bm A$, where $\bm{A}\defeq\sum_{i=1}^{M}\bm{H}_{i}\tr\boldsymbol{H}_{i}\defeq\left[\begin{array}{cc}
\boldsymbol{A}_{rr} & \boldsymbol{A}_{rt}\\
\boldsymbol{A}_{rt}\tr & \boldsymbol{A}_{tt}
\end{array}\right]$, $\bm H_i\defeq [-(\bm p_i\times \bm n_i)^\top,\ -\bm n_i^\top]$, and  $\bm{n}_i$ is the surface unit normal vector based at a point $\bm{p}_i$.
The eigenvector associated with the smallest eigenvalue of $\bm A$ is the \textit{least observable} direction for pose estimation. The translational part is usually the most challenging part of the pose to estimate, mainly because of the presence of long shafts and corridors. 
We propose therefore to use the condition number $\kappa(\boldsymbol{A}_{tt}):=\frac{\left|\lambda_{max}(\boldsymbol{A}_{tt})\right|}{\left|\lambda_{min}(\boldsymbol{A}_{tt})\right|}$ as the observability metric. The larger the condition number $\kappa(\boldsymbol{A}_{tt})$ is, the more poorly constrained the optimization problem is in the translational part. When this condition number is above a user-defined threshold, LION issues a warning to the switching logic HeRO \cite{santamaria2019towards}, so that other more reliable odometry sources can be used. 

\vspace{-0.5cm}
\section{EXPERIMENTAL RESULTS}\label{sec:results}

\subsection{LION estimation performance in the 2019 Tunnel competition}
We first evaluate the performance of LION (translation and rotation errors, and repeatability of the results) in the two runs of the two different tracks of the Tunnel Competition, held in the NIOSH experimental mines in Pittsburgh, USA. 
Lidar odometry was computed at 10 Hz, while IMU and LION output could be provided at up to 200 Hz. The sliding window of LION used is 3 seconds and LION back-end was tuned to use approximately 30\% of one CPU core of an i7 Intel NUC.
For reference, we compare its performance with  Wheel-Inertial odometry (wheel odometry fused with an IMU via an Extended Kalman Filter) and Scan-To-Scan odometry (the relative pose input of LION), with LAMP~\cite{Ebadi20lamp} as ground truth.

\begin{figure}[!tbp]
  \centering
  \begin{minipage}[b]{0.48\textwidth}
    \includegraphics[trim=0 250 50 120,clip,width=\textwidth]{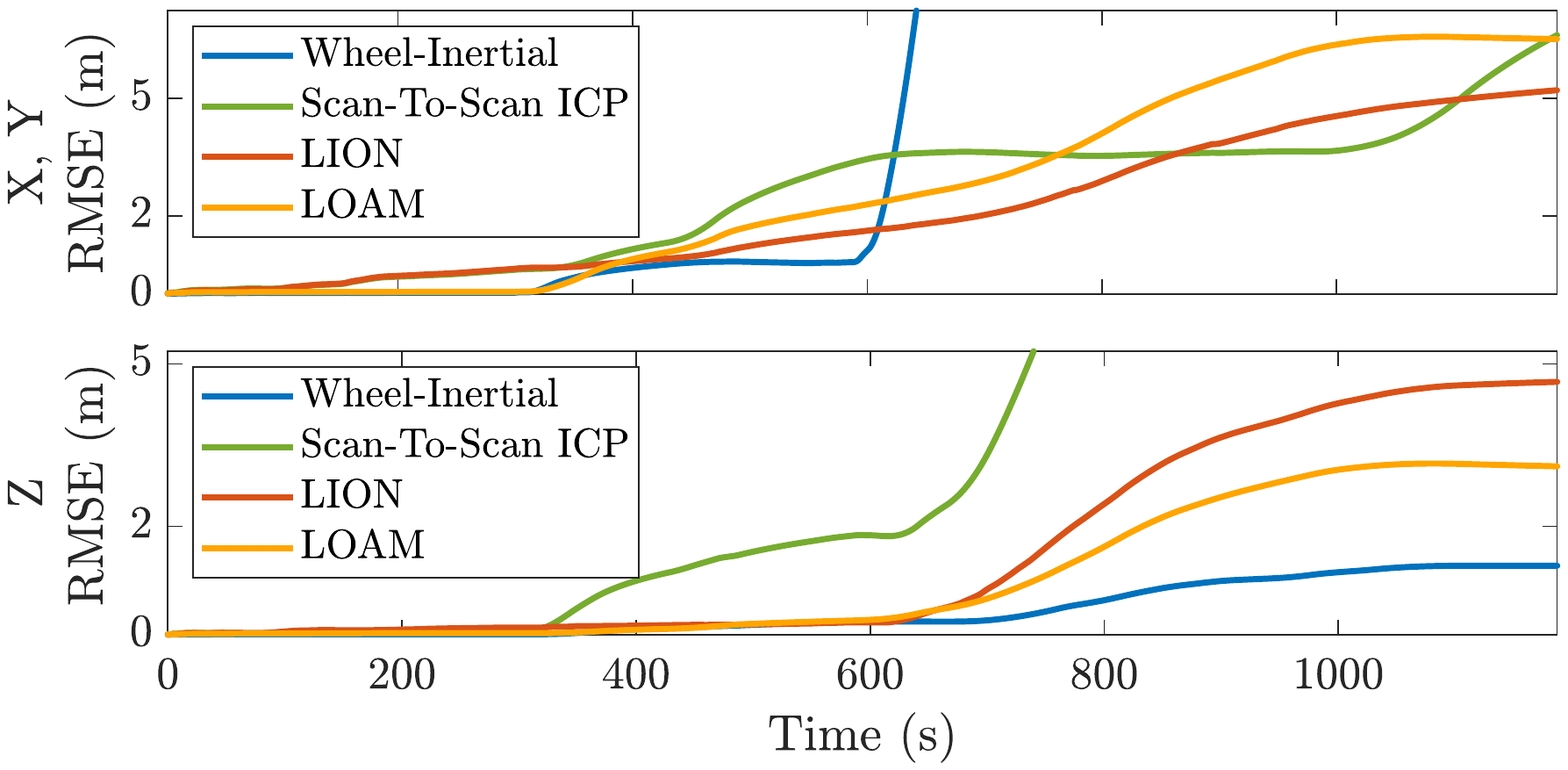}
    \caption{Position estimation RMSE. (Track A, Run 2)}
    \label{fig:day1_h1_position_error}
    \vspace*{-.5in}
  \end{minipage}
  \hfill
  \begin{minipage}[b]{0.48\textwidth}
    \includegraphics[trim=0 250 50 120, clip,width=\textwidth]{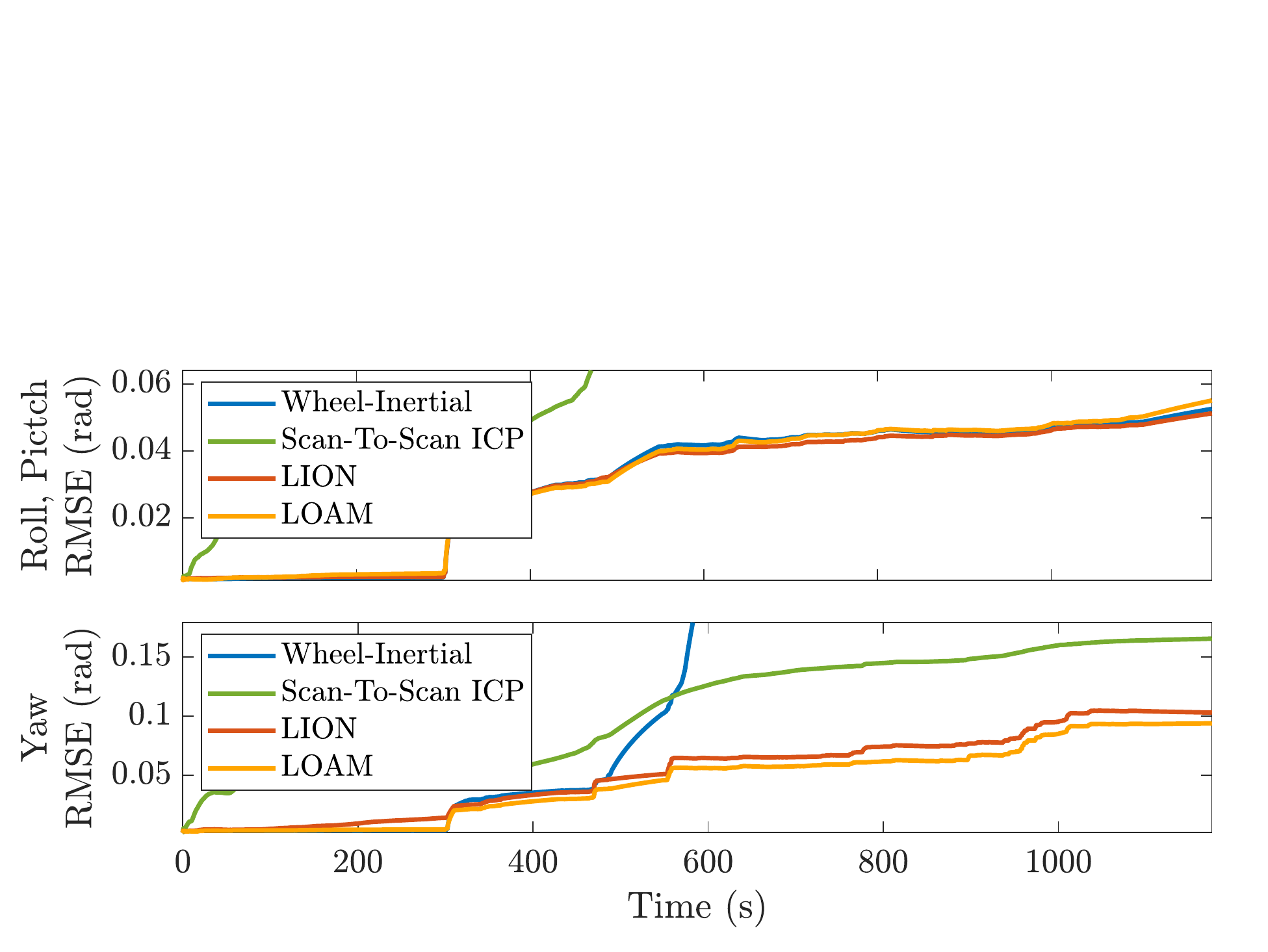}
    \caption{Attitude estimation RMSE. (Track A, Run 2)}
    \label{fig:day1_h1_attitude_error}
    \vspace*{-.5in}
  \end{minipage}
  \vspace*{-.05in}
    \centering
    \begin{minipage}[b]{0.48\textwidth}
    \includegraphics[trim=30 265 40 130,clip,width=1\textwidth]{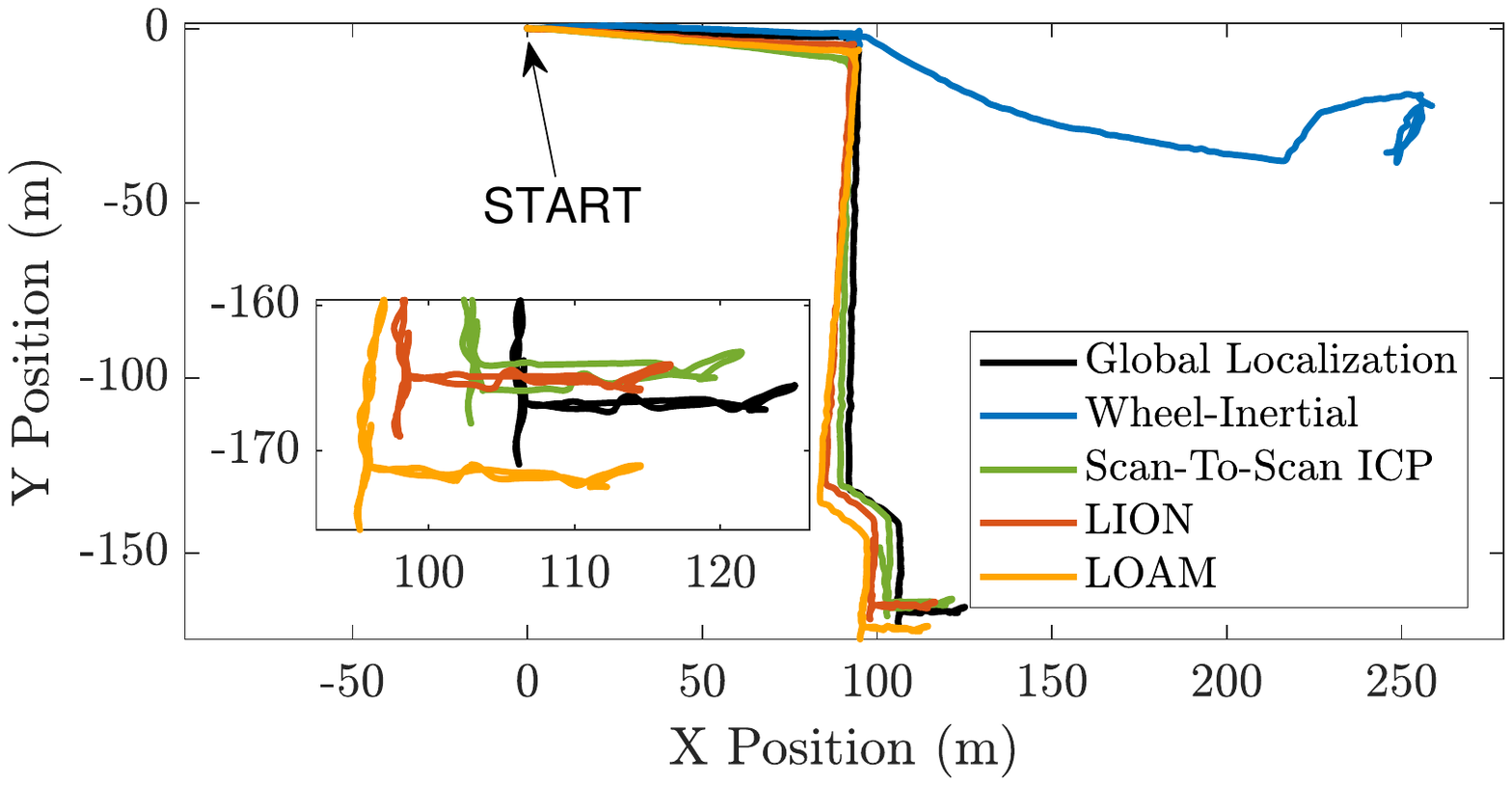}
    \caption{$x,y$ position estimates. (Track A, Run 2, first 1000 s)}
    \label{fig:day1_h1_XYZ}
  \end{minipage}
  \hfill
  \begin{minipage}[b]{0.48\textwidth}
    \includegraphics[trim=0 0 0 0, clip, width=\textwidth]{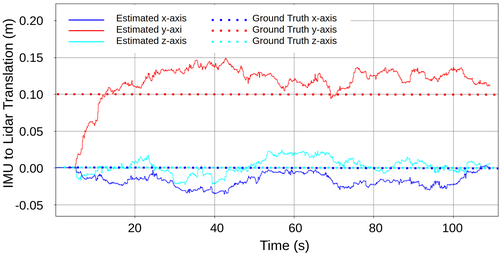}
    \caption{Estimation of the translation of the extrinsics.}
    \label{fig:auto_calibration_error}
    
  \end{minipage}
  \vspace*{-.05in}
\end{figure}

The results are summarized in Table \ref{table:tunnel_stats}. We can see that fusing inertial data with the odometry from the front-end (LION) significantly reduces the drift of the pure lidar-odometry method (Scan-To-Scan). This is especially evident from the estimation error along the $z$-axis shown in Fig.~\ref{fig:day1_h1_position_error}. Additionally, LION reliably estimates the attitude of the robot (Fig.~\ref{fig:day1_h1_attitude_error}), achieving small roll and pitch errors (since they are observable in the IMU via the gravity vector) and yaw. Fig.~\ref{fig:day1_h1_XYZ} shows that LION outperforms the baseline approach (Wheel-Inertial) especially in terms of drift in yaw and horizontal translation. To showcase LION's auto-calibration capabilities, we generate in simulation a dataset where the lidar is translated of $0.1$ m with respect to the IMU, along the IMU's $y$-axis. In Fig.~\ref{fig:auto_calibration_error} we observe that after approximately $20$ s, LION (continuous lines) estimates the correct extrinsics (dotted lines). 
Last, we compare  LION with the state-of-the-art method LOAM \cite{zhang2014loam}. The comparison, shown in Fig.~\ref{fig:day1_h1_position_error}, \ref{fig:day1_h1_attitude_error} and \ref{fig:day1_h1_XYZ} and Table \ref{table:tunnel_stats}, highlights that LION performs comparably to LOAM for short trajectories ($t < 600$ s in Fig~\ref{fig:day1_h1_position_error}, and Table~\ref{table:tunnel_stats}, Track B, Run 2), while LOAM (thanks to the presence of a map which compensates for drifts) achieves lower position and yaw drift for longer trajectories. LION obtains comparable or slightly lower roll and pitch errors thanks to the fusion of IMU data (Fig. \ref{fig:day1_h1_attitude_error}), which guarantees a gravity-aligned output provided at IMU rate. The output rate of LOAM  is instead limited to the rate of the lidar. 

\begin{table}
\begin{centering}
\resizebox{\columnwidth}{!}{
\begin{tabular}{|c||c|c|c|c|c|c||c|c|c|c|c|c||}
\cline{2-13} \cline{3-13} \cline{4-13} \cline{5-13} \cline{6-13} \cline{7-13} \cline{8-13} \cline{9-13} \cline{10-13} \cline{11-13} \cline{12-13} \cline{13-13} 
\multicolumn{1}{c||}{} & \multicolumn{6}{c||}{\textbf{Track A}} & \multicolumn{6}{c||}{\textbf{Track B}}\tabularnewline
\hline 
 & \multicolumn{3}{c|}{\textbf{Run 1} (685 m, 1520 s)} & \multicolumn{3}{c||}{\textbf{Run 2} (456 m, 1190 s)} & \multicolumn{3}{c|}{\textbf{Run 1 }(467 m, 1452 s)} & \multicolumn{3}{c||}{\textbf{Run 2 }(71 m, 246 s)}\tabularnewline
\hline 
\hline 
\textbf{Algorithm} & $\boldsymbol{t}$(m) & $\boldsymbol{t}$(\%) & $\boldsymbol{R}$(rad) & $\boldsymbol{t}$(m) & $\boldsymbol{t}$(\%) & $\boldsymbol{R}$(rad) & $\boldsymbol{t}$(m) & $\boldsymbol{t}$(\%) & $\boldsymbol{R}$(rad) & $\boldsymbol{t}$(m) & $\boldsymbol{t}$(\%) & $\boldsymbol{R}$(rad)\tabularnewline
\hline 
\textbf{Wheel-Inertial}& 130.50&	19.05&	1.60&	114.00&	25.00&	1.28&	78.21&	16.75&	0.99&	6.91&	9.79&   0.12\tabularnewline
\hline 
\textbf{Scan-To-Scan} & 105.47&	15.40&	0.90&	18.72&	4.11&	0.18&	56.6&	12.14&	0.79&	4.55&	6.45&	0.27\tabularnewline
\hline 
\textbf{LION} & 56.92&	8.31&	0.36&	7.00&	1.53&	0.10&	17.59&	3.77&	0.27&	3.78&	5.36&	0.05\tabularnewline
\hline 
\textbf{LOAM} & 10.99&	    1.60&	0.14&	7.22&	1.58&	0.08&	13.21&	2.83&	0.21&	5.55&	7.87&	0.03\tabularnewline
\hline
\end{tabular}
} %
\par\end{centering}
\caption{Estimation error of Wheel-Inertial Odometry, Scan-To-Scan Matching, LION, and LOAM for two runs of the two tracks of the Tunnel competition, computed for one of the robots deployed. 
The second row shows the total distance and time traveled in each run. Note $\boldsymbol{t}$(m) and $\boldsymbol{R}$(rad) indicate the RMSE for position and attitude estimation, and $\boldsymbol{t}$(\%) indicates the percentage drift in position.}
\label{table:tunnel_stats}
 \vspace*{-.15in}
\end{table}

\begin{figure}
    \centering
    \begin{minipage}[b]{1.0\textwidth}
    \includegraphics[trim=0 0 0 0,clip,width=\textwidth]{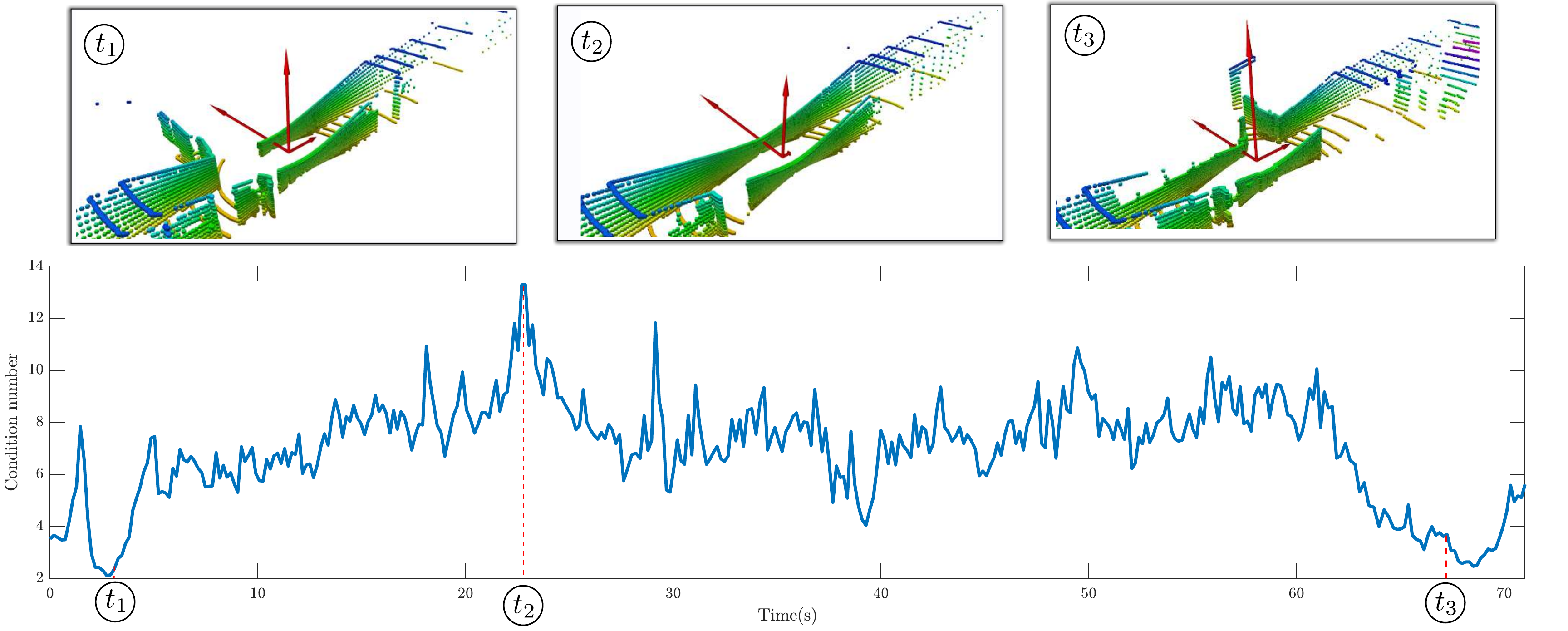}
    \caption{Plot of the condition number $\kappa(\bm A_{tt})$ in the \texttt{JPL-Corridor} dataset (bottom), together with three snapshots of the eigenvectors of $\bm A_{tt}$ scaled with their corresponding eigenvalues (top).}
    \label{fig:cond_number_corridor}
  \end{minipage}
\end{figure}
\begin{figure}
  \begin{minipage}[b]{1.0\textwidth}
    \includegraphics[trim=0 0 0 0, clip, width=\textwidth]{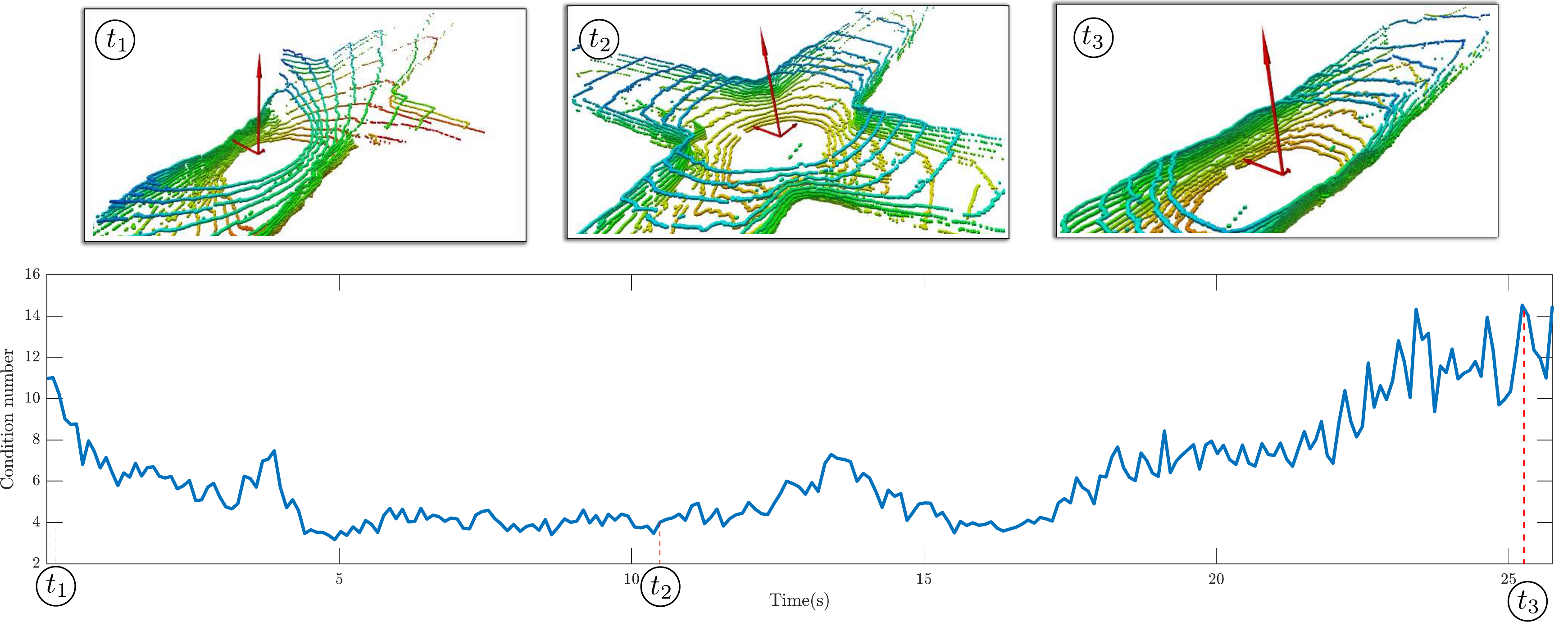}
    \caption{Plot of the condition number $\kappa(\bm A_{tt})$ in the \texttt{Arch-Coal-Mine} dataset (bottom), together with three snapshots of the eigenvectors of $\bm A_{tt}$ scaled with their corresponding eigenvalues (top)}
    \label{fig:cond_number_tunnel}
  \end{minipage}
   \vspace*{-.15in}
\end{figure}

\subsection{Observability module}
We first show how the condition number can detect geometrically-unconstrained scenes. To test this and to build intuition, we first use the dataset \texttt{JPL-Corridor}, recorded in an office at JPL, whose main challenge is the lack of geometric features along the direction of the corridor. We also use the dataset \texttt{Arch-Coal-Mine}, recorded in the Arch Coal Mine (see Fig. \ref{fig:cond_number_corridor}), which consists of a straight tunnel followed by an intersection. The plots of the condition number and the eigenvectors scaled with the eigenvalues for the \texttt{JPL-Corridor} dataset are shown in Fig. \ref{fig:cond_number_corridor}. At the beginning and the end of the corridor, there are enough features in all the directions, and the condition number is $\kappa(\bm A_{tt})\approx 2$. However, in the middle of the corridor, the condition number reaches values $\kappa(\bm A_{tt}) > 13$, which happens when the eigenvalue associated with the eigenvector along the direction of the tunnel becomes really small. This big condition number makes it hard %
to determine if the robot is moving based on changes in the geometry of the incoming point clouds. A similar situation happens in the \texttt{Arch-Coal-Mine} dataset (see Fig. \ref{fig:cond_number_tunnel}). Before and after the intersection, $\kappa(\bm A_{tt}) > 10$ indicates the low observability along the main shaft. The observability is 
improved %
when the robot reaches the intersection, in which the condition number is $\kappa(\bm A_{tt}) \approx 4$.  Note also the symmetry in the intersection of the two scaled eigenvectors that point to the two different corridors. 

\begin{figure*}[t]
\centering
\includegraphics[width=\linewidth]{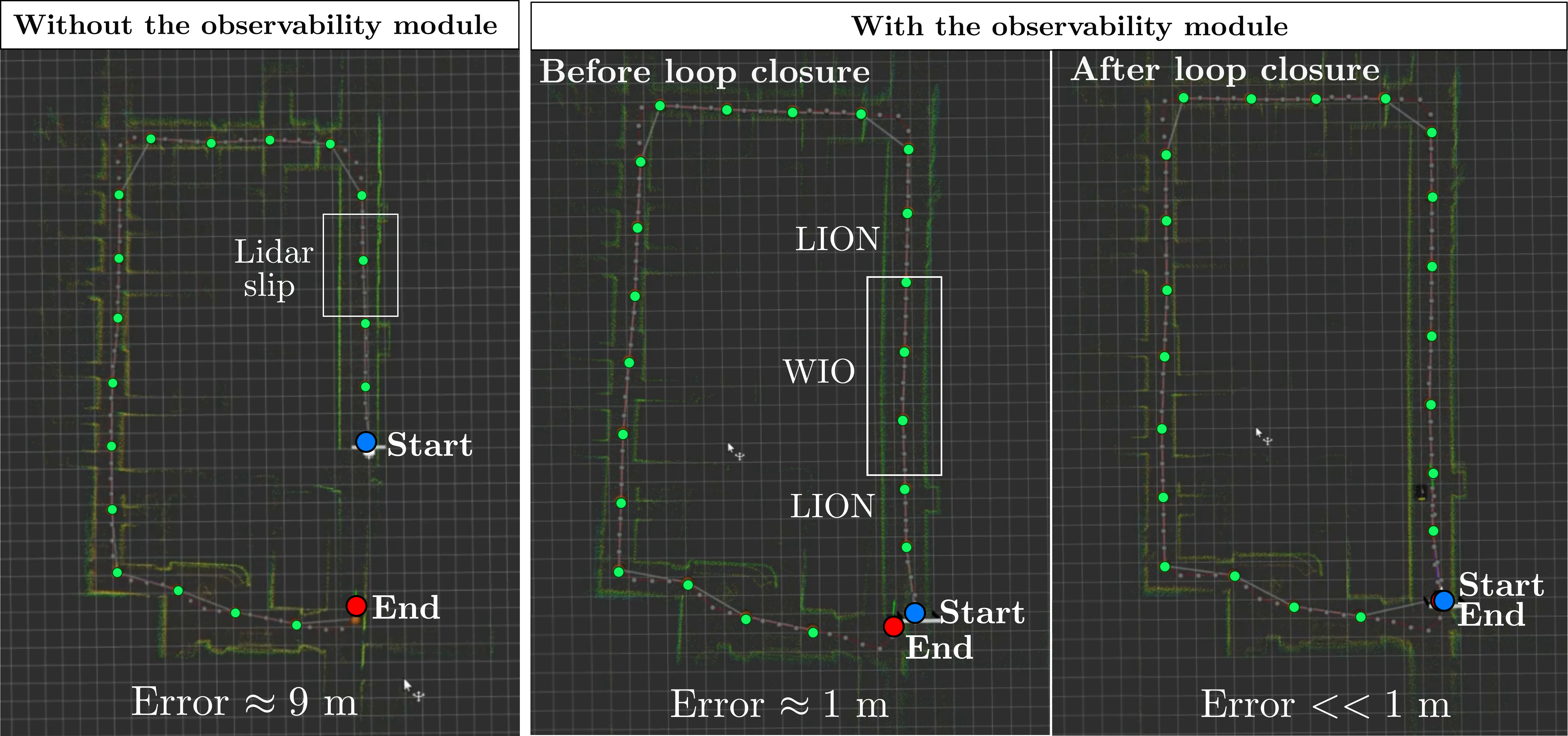}
  	\caption{Comparison of the translation error with and without the observability module in an office-like environment. On the left, the observability module is not used, which creates a \textit{lidar slip} in the first part of the corridor, producing an error of $\approx 9$ m. On the right, the observability module is used, and the switching-logic HERO switches to WIO instead of LION for the section of the corridor without lidar features. The total error is $\approx 1$ m (``Before loop closure''). Improved state estimation (reduced drift in the output of HeRO) benefits the global mapping solution \cite{Ebadi20lamp}, which can now correctly detect a loop closure (Fig.~\ref{fig:slip_office}  ``After loop closure''), further reducing the drift. \vspace{-5pt}}
    \label{fig:slip_office}
    \vspace{-10pt}
\end{figure*}

Using this condition number as the observability metric, HeRO can decide to switch to other odometry sources different from LION (like WIO) when there are not enough lidar features. %
This behavior is shown in Fig.~\ref{fig:slip_office} for a real experiment carried out in an office-like environment where a section of the first corridor does not have enough lidar features. If the observability module is not used, this lack of features in the corridor creates a \textit{lidar slip}, where the movement of the robot is not well observed by LION. This produces a total error of $\approx 9$ m when the robot goes back to the original position. However, when the observability module is used, the lack of observability in the direction of the corridor is detected, and WIO is used for that section of the corridor, instead of LION. This produces a total error of $\approx 1$ m, which is small enough to trigger a successful loop-closure in the employed global mapping solution \cite{Ebadi20lamp}.

\vspace{-1.0cm}

\section{EXPERIMENTAL INSIGHTS AND CONCLUSIONS}\label{sec:conc}
This work presented LION, a Lidar-Inertial Observability-Aware algorithm for Navigation of robots in vision-denied environments.
LION has been extensively tested in different subterranean environments, including the Tunnel Competition of the DARPA Subterrranean Challenge in August 2019. 
In the following we review the insights and reasons behind some of the choices made, as well as the main outstanding technical challenges.

\begin{itemize}%
    \item \textbf{Local state estimator}: The main goal of LION is to provide a high-rate, continuous and smooth output to the downstream algorithms. As such, LION does not build any map and does not perform any loop closures, and thus its inertial reference frame will slowly drift over time. The slow drifts in the output of LION are compensated by the mapping strategy LAMP \cite{Ebadi20lamp}, which is running within the localization framework. %
    \item \textbf{Loosely-coupled architecture}: In contrast to other state-of-the-art works \cite{shan2018lego, ye2019tightly, zhang2014loam}, the lidar front-end and back-end of LION are loosely-coupled (i.e., we do not explicitly add features point or scans in the state of the estimator). The main reason behind this was the desire of sharing the computational resources required by the lidar front-end with the mapping solution \cite{Ebadi20lamp}. In addition, such architecture was dictated by the need of modularity in choosing front-end/back-end algorithms, and distributing the risk between several estimation engines to remove the one-point failure in case of a single estimation engine. %
    \item \textbf{Not feature-based}: In LION, and contrary to \cite{zhang2014loam}, no features are extracted to match two point clouds because of two reasons. On one hand, feature extraction is usually computationally expensive, which would lead to a reduction of the performance of other modules of LION due to the limited computational resources onboard. Moreover, LION is meant for exploration of completely unknown environments, where there could be human-made structures (full of corners, planes and lines) or completely unstructured terrains. The use of feature extraction for such uncertain environments poses a prior on the environment the robot is going to navigate through, leading therefore to its own risks or failures. %
    \item \textbf{Automatic extrinsic calibration}: lidar to IMU calibration is critical, especially in the relative orientation of the two sensors, since small errors can quickly integrate and cause a large drift. Offline calibration methods require a calibration target \cite{le20183d} and/or specific motion sequences \cite{della2019unified}, and can be impractical in a field setup where the sensor can be re-positioned at the last minute. This is the reason why we choose to estimate the extrinsics directly with LION. 
    \item  \textbf{Supervisory Algorithm}: LION has been designed to be one of the many odometry sources used by HeRO \cite{santamaria2019towards}, a switching logic
    that multiplexes different odometry sources such as wheel encoders, visual-inertial or thermal-inertial odometry. The selection of one of these odometry sources is done according to the reliability of their output as state estimate. The confidence metrics for LION include its output rate, the detection of the potential failures in its inputs (lidar and IMU), and the observability score detailed in Section \ref{sec:observability}.
\end{itemize}

\section*{ACKNOWLEDGMENTS}
The authors would like to thank Kasra Khosoussi (ACL-MIT), Benjamin Morrell (JPL) and Kamak Ebadi (JPL) for helpful insights and discussions. Part of this  research was carried out at the Jet Propulsion Laboratory, California Institute of Technology, under a contract with the National Aeronautics and Space Administration.

 \vspace*{-.3in}

\bibliographystyle{IEEEtran}

{\tiny \bibliography{bibliography.bib} }

\begin{acronym}
\acro{LION}{Lidar-Inertial Observability-Aware Navigator}
\acro{VIO}{Visual-Inertial Odometry}
\acro{TIO}{Thermal-Inertial Odometry}
\acro{WO}{Wheel-Odometry}
\end{acronym}

\end{document}